\documentclass[11pt,a4paper]{article}
\usepackage{times}
\usepackage{latexsym}

\usepackage{graphicx}
\usepackage[small]{caption}
\usepackage{subcaption}
\usepackage{natbib}
\usepackage{appendix}

\usepackage[utf8]{inputenc}
\usepackage[T1]{fontenc}
\usepackage{hyperref}
\usepackage{url}
\usepackage{booktabs}
\usepackage{amsfonts}
\usepackage{nicefrac}
\usepackage[english]{babel}
\usepackage[activate=true,
    final,
    tracking=false,
    kerning=true,
    spacing=true,
    factor=1000,
    stretch=5,
    shrink=40]{microtype}
\usepackage{xpatch}
\xapptocmd\normalsize{
 \abovedisplayskip=11pt plus 3pt minus 9pt
 \abovedisplayshortskip=0pt plus 3pt
 \belowdisplayskip=11pt plus 3pt minus 9pt
 \belowdisplayshortskip=6.5pt plus 3.5pt minus 3pt
}{}{}

\usepackage[]{comment}
\usepackage{algorithm}
\usepackage{algorithmic}
\usepackage{amsmath}
\usepackage{amssymb}
\usepackage{mathtools}
\usepackage{xspace}

\usepackage[noabbrev,capitalize]{cleveref}

\crefname{equation}{equation}{equations}
\crefname{footnote}{footnote}{footnotes}
\crefname{section}{\S}{\S\S}
\Crefname{section}{\S}{\S\S}
\crefformat{section}{#2\S#1#3}
\Crefformat{section}{#2\S#1#3}
\crefrangeformat{section}{\S\S#3#1#4--#5#2#6}
\Crefrangeformat{section}{\S\S#3#1#4--#5#2#6}

\usepackage[usenames,dvipsnames,svgnames,table]{xcolor}
\usepackage[disable]{todonotes}

\newcommand{\cutforspace}[1]{}

\newcommand{\defeq}{\mathrel{\stackrel{\mbox{\tiny def}}{=}}}
\newcommand{\defpropto}{\mathrel{\stackrel{\mbox{\tiny def}}{\propto}}}
\usepackage{bm}
\newcommand{\vecc}[1]{\boldsymbol{\mathbf{#1}}}

\newcommand{\vx}{{\vecc{x}}}
\newcommand{\vy}{{\vecc{y}}}
\newcommand{\vs}{{\vecc{s}}}
\newcommand{\vsbar}{{\bar{\vs}}}
\newcommand{\vz}{{\vecc{z}}}

\newcommand{\vu}{{\vecc{u}}}
\newcommand{\valpha}{\vecc{\alpha}}
\newcommand{\vbeta}{\vecc{\beta}}
\newcommand{\vone}{\mathbf{1}}
\newcommand{\overbar}[1]{\mkern 1.5mu\overline{\mkern-1.5mu#1\mkern-1.5mu}\mkern 1.5mu}
\newcommand{\Pbar}{{\overbar{P}}}
\newcommand{\phat}{{\hat{p}}}
\newcommand{\bos}{\textsc{bos}\xspace}
\newcommand{\eos}{\textsc{eos}\xspace}
\DeclareMathOperator*{\argmin}{argmin}

\usepackage{stmaryrd}
\newcommand{\norm}[1]{\llbracket #1 \rrbracket}
\newcommand{\psup}[1]{^{(#1)}}

\newcommand{\defn}[1]{\textbf{#1}}
\newcommand{\Real}{\mathbb{R}}
\newcommand{\E}[2][]{\mathbb{E}_{{#1}}\left[#2\right]}
\usepackage[hyperref]{naaclhlt2018}
\usepackage{url}
\aclfinalcopy

\title{Neural Particle Smoothing \\ for Sampling from Conditional Sequence Models}
\date{}
\begin{document}
\author{Chu-Cheng Lin \and Jason Eisner \\ Center for Language and Speech Processing \\ Johns Hopkins University, Baltimore MD, 21218 \\ \texttt{\{kitsing,jason\}@cs.jhu.edu}}
\maketitle
\begin{abstract}
We introduce \emph{neural particle smoothing}, a sequential Monte Carlo method for sampling annotations of an input string from a given probability model.  In contrast to conventional particle filtering algorithms, we train a  proposal distribution that {\em looks ahead} to the end of the input string by means of a right-to-left LSTM.  We demonstrate that this innovation can improve the quality of the sample.  To motivate our formal choices, we explain how
  our neural model and neural sampler can be viewed as
  low-dimensional but nonlinear approximations to working with HMMs over very large state spaces.\looseness=-1
\end{abstract}
\section{Introduction}
\label{sec:intro}
Many structured prediction problems in NLP can be reduced to labeling a
length-$T$ input string $\vx$ with a length-$T$ sequence $\vy$ of
tags.
In some cases, these tags are annotations such as syntactic parts of
speech.  In other cases, they represent actions that incrementally
build an output structure: IOB tags build a chunking of the input \cite{ramshaw1999}, shift-reduce actions
build a tree \cite{yamada2003}, and finite-state transducer arcs build an output
string \cite{pereira1997}.%

One may wish to score the possible taggings using a recurrent neural network, which can learn to be sensitive to complex patterns in the training data.  A globally normalized conditional probability model is particularly valuable because it quantifies uncertainty and does not suffer from label bias \cite{lafferty-mccallum-pereira-2001}; also, such models often arise as the predictive conditional distribution $p(\vy\mid\vx)$ corresponding to some well-designed generative model $p(\vx,\vy)$ for the domain.  In the neural case, however, inference in such models becomes intractable.  It is hard to know what the model actually predicts and hard to compute gradients to improve its predictions.

In such intractable settings, one generally falls back on approximate inference or sampling.  In the NLP community, beam search and importance sampling are common.  Unfortunately, beam search considers only the approximate-top-$k$ taggings from an exponential set \cite{wiseman2016}, and importance sampling requires the construction of a good proposal distribution \cite{dyer2016}.

In this paper we exploit the sequential structure of the tagging problem to do {\em sequential} importance sampling, which resembles beam search in that it constructs its proposed samples incrementally---one tag at a time, taking the actual model into account at every step.  This method is known as particle filtering \cite{doucet2009}.  We extend it here to take advantage of the fact that the sampler has access to the entire input string as it constructs its tagging, which allows it to look ahead or---as we will show---to use a neural network to approximate the effect of lookahead.  Our resulting method is called {\em neural particle smoothing}.

\subsection{What this paper provides}\label{sec:setup}

For $\vx = x_1 \cdots x_T$, let $\vx_{:t}$ and $\vx_{t:}$ respectively denote the prefix $x_1 \cdots x_t$ and the suffix $x_{t+1} \cdots x_T$.

We develop \emph{neural particle smoothing}---a sequential importance
sampling  method which, given a string $\vx$, draws a sample of taggings $\vy$ from
$p_\theta(\vy \mid \vx)$.  Our method works
for any conditional probability model of the quite general form\footnote{\label{fn:eos}A model may require for convenience
  that each input end with a special end-of-sequence symbol: that is,
  $x_T = \eos$.}
\begin{align}\label{eq:condit}
p_\theta(\vy \mid \vx) &\defpropto \exp G_T
\end{align}
where $G$ is an \emph{incremental stateful global scoring model} that
recursively defines scores $G_t$ of prefixes of $(\vx,\vy)$ at all times
$0 \leq t \leq T$:
\begin{align}
\!\!\!\!G_t &\defeq G_{t-1} + g_\theta(\vs_{t-1},x_t,y_t) &\!\!\!\!\text{(with $G_0\defeq 0$)} \label{eq:G} \\
\!\!\!\!\vs_t &\defeq f_\theta(\vs_{t-1},x_t,y_t) &\!\!\!\!\text{(with $\vs_0$ given)} \label{eq:s}
\end{align}

These quantities implicitly depend on $\vx,\vy$ and $\theta$.
Here $\vs_t$ is the model's \emph{state} after
observing the pair of length-$t$ prefixes $(\vx_{:t},\vy_{:t})$.
$G_t$ is the \emph{score-so-far} of this prefix pair, while
$G_T - G_t$ is the \emph{score-to-go}.  The state $\vs_t$ summarizes
the prefix pair in the sense that the score-to-go
depends only on $\vs_t$ and the length-$(T-t)$ suffixes $(\vx_{t:}, \vy_{t:})$.
The \emph{local scoring function} $g_\theta$ and \emph{state update
  function} $f_\theta$ may be any functions parameterized by
$\theta$---perhaps neural networks.
We assume $\theta$ is fixed and given.

This model family is expressive enough to capture any desired
$p(\vy\mid \vx)$.  Why?  Take any distribution $p(\vx,\vy)$ with this
desired conditionalization (e.g., the true joint distribution) and
factor it as
\begin{align}
\!\!\!\!  \log p(\vx,\vy)\!
  &= \textstyle \sum_{t=1}^{T} \log p(x_t,y_t \mid \vx_{:t-1},\vy_{:t-1}) \nonumber \\
  &= \textstyle \sum_{t=1}^{T} \underbrace{\log p(x_t,y_t \mid
    \vs_{t-1})}_{\text{use as }g_\theta(\vs_{t-1},x_t,y_t)} = G_T \label{eq:noindep}
\end{align}
by making $\vs_t$ include as much information about
$(\vx_{:t},\vy_{:t})$ as needed for
\eqref{eq:noindep} to hold
(possibly $\vs_t=(\vx_{:t},\vy_{:t})$).\footnote{Furthermore, $\vs_t$
  could even depend on all of $\vx$ (if $\vs_0$ does), allowing direct expression
  of models such as stacked BiRNNs.}
  Then by defining $g_\theta$ as shown in \eqref{eq:noindep},
we get $p(\vx,\vy) = \exp G_T$ and thus \eqref{eq:condit} holds for
each $\vx$.

\subsection{Relationship to particle filtering}\label{sec:Hhat}

Our method is spelled out in \cref{sec:sis} (one may look now). It is a variant of the popular \emph{particle filtering} method that tracks the state of a physical system in discrete time \cite{ristic2004}.  Our particular \emph{proposal distribution} for $y_t$ can be found in \cref{eq:Hhat,eq:sbar,eq:c,eq:q-local}. It considers not only past observations $\vx_{:t}$ as reflected in $\vs_{t-1}$, but also future observations $\vx_{t:}$, as summarized by the state $\vsbar_t$ of a right-to-left recurrent neural network $\bar{f}$ that we will train:
\begin{align}
\hat{H}_t &\defeq h_\phi(\vsbar_{t+1},x_{t+1}) + \hat{H}_{t+1}\label{eq:Hhat} \\
\vsbar_t &\defeq \bar{f}_\phi(\vsbar_{t+1},x_{t+1}) &\!\!\!\!\!\!\!\!\text{(with $\vs_T$ given)}\label{eq:sbar}
\end{align}
Conditioning the distribution of $y_t$ on future observations $\vx_{t:}$ means that we are
doing ``smoothing'' rather than ``filtering'' (in signal processing terminology).  Doing so can reduce the bias and variance of our sampler.  It is possible so long as $\vx$ is provided in its entirety before the sampler runs---which is often the case in NLP.

\begin{figure*}[t]
\centering
\includegraphics[width=.93\textwidth]{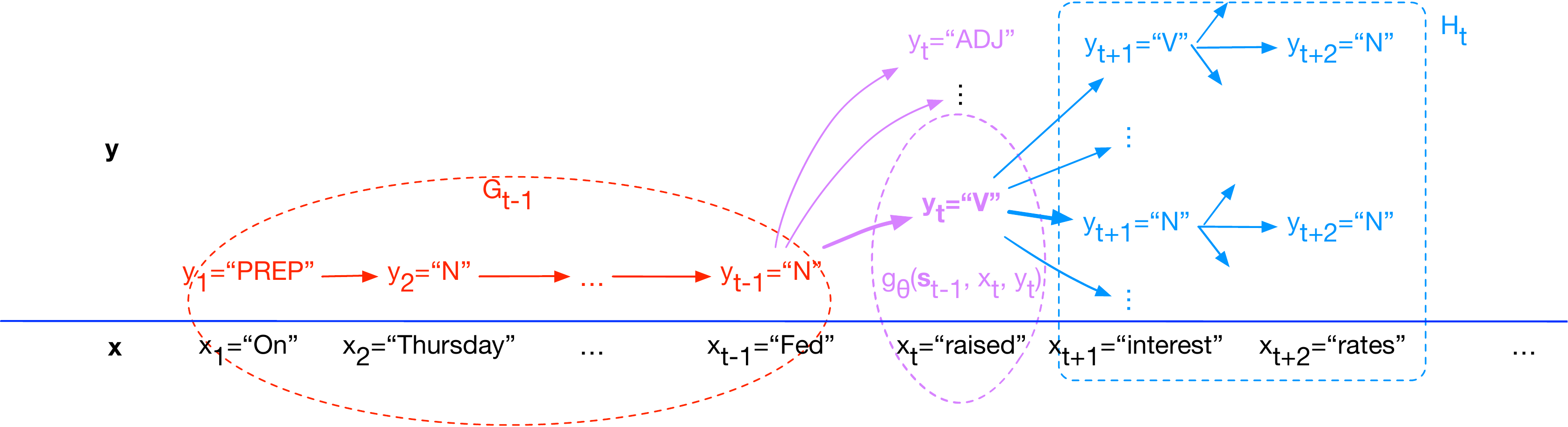}
\caption{\label{fig:quantities} Sampling a single particle from a tagging model.  $y_1,\ldots,y_{t-1}$ (orange) have already been chosen, with a total model score of $G_{t-1}$, and now the sampler is constructing a proposal distribution $q$ (purple) from which the next tag $y_t$ will be sampled.  Each $y_t$ is evaluated according to its contribution to $G_t$ (namely $g_\theta$) and its future score $H_t$ (blue).  The figure illustrates quantities used throughout the paper, beginning with exact sampling in \crefrange{eq:q-exact}{eq:q-exact-unnorm}.  Our main idea (\cref{sec:approx}) is to {\em approximate} the $H_t$ computation (a log-sum-exp over exponentially many sequences) when exact computation by dynamic programming is not an option.  The form of our approximation uses a right-to-left recurrent neural network but is {\em inspired} by the exact dynamic programming algorithm.
  \cutforspace{At time step $t$, the sequential sampler has accumulated the logprob-so-far in $G_{t-1}$. The distribution over the next tag $p_\theta(y_t \mid \vx, \vy_{:t-1})$ is proportional to $\exp (g_{\theta}(\vs_{t-1}, x_t, y_t) + H_t)$. If we could compute $H_t$ exactly, we would be able to sample from \cref{eq:q-exact} exactly when the set of possible tags $\mathcal{Y}$ is not too large. However, in the general case there is no reusable computation within $H_t$ (no states are being shared), making exact computation intractable.}}
\end{figure*}

\subsection{Applications}\label{sec:apps}

Why sample from $p_\theta$ at all?  Many NLP systems instead simply search for the \emph{Viterbi sequence} $\vy$ that maximizes $G_T$ and thus maximizes $p_\theta(\vy \mid \vx)$.  If the space of states $\vs$ is small, this can be done efficiently by dynamic programming \cite{viterbi-1967}; if not, then $A^*$ may be an option (see \cref{sec:astar}).  More common is to use an approximate method: beam search, or perhaps a sequential prediction policy trained with reinforcement learning.  Past work has already shown how to improve these approximate search algorithms by conditioning on the future \cite{bahdanau2016,wiseman2016}.

Sampling is essentially a generalization of maximization: sampling from $\exp \frac{G_T}{\mathrm{temperature}}$ approaches maximization as $\mathrm{temperature} \rightarrow 0$.  It is a fundamental building block for other algorithms, as it can be used to take expectations over the whole space of possible $\vy$ values.  For unfamiliar readers, \cref{app:apps} reviews how sampling is crucially used in minimum-risk decoding, supervised training, unsupervised training, imputation of missing data, pipeline decoding, and inference in graphical models.\looseness=-1

\section{Exact Sequential Sampling}
\label{sec:exact}
\label{sec:astar}

To develop our method, it is useful to first consider exact samplers.
Exact sampling is tractable for only some of the models allowed by \cref{sec:setup}.  However, the form and notation of the exact algorithms in \cref{sec:exact} will guide our development of approximations in \cref{sec:approx}.

An \emph{exact sequential sampler} draws $y_t$ from $p_\theta(y_t \mid \vx,\vy_{:t-1})$ for each $t = 1, \cutforspace{2,} \ldots, T$ in sequence.  Then $\vy$ is exactly distributed \cutforspace{according to}{as} $p_\theta(\vy \mid \vx)$.

For each given $\vx, \vy_{:t-1}$, observe that
\begin{align}
\makebox[1.5em][l]{$\displaystyle p_\theta(y_t \mid \vx,\vy_{:t-1})$} \label{eq:q-exact} \\
  & \propto p_\theta(\vy_{:t} \mid \vx)
    = \textstyle\sum_{\vy_{t:}} p_\theta(\vy \mid \vx) \\
\quad
  & \propto \textstyle \sum_{\vy_{t:}} \exp G_T \label{eq:GplusH} \\
  &= \exp\;(G_t + \underbrace{\log \textstyle \sum_{\vy_{t:}} \!\exp\;(G_T - G_t)}_{\text{call this $H_t$}})\!\! \label{eq:H} \\
  &= \exp\; (G_{t-1} + g_\theta(\vs_{t-1},x_t,y_t) + H_t) \\
  &\propto \exp\;(g_\theta(\vs_{t-1},x_t,y_t) + H_t) \label{eq:q-exact-unnorm}
\end{align}
Thus, we can easily construct the needed distribution \eqref{eq:q-exact} by normalizing \eqref{eq:q-exact-unnorm} over all possible values of $y_t$.
The challenging part of \eqref{eq:q-exact-unnorm} is to compute $H_t$: as
defined in \eqref{eq:H}, $H_t$ involves a sum over exponentially many futures $\vy_{t:}$. (See \cref{fig:quantities}.)

We chose the symbols $G$ and $H$ in homage to the $A^*$ search algorithm \cite{ASTAR-1968}.  In that algorithm (which could be used to find the Viterbi sequence), $g$ denotes the score-so-far of a partial solution $\vy_{:t}$, and $h$ denotes the optimal score-to-go.  Thus, $g+h$ would be the score of the \emph{best} sequence with prefix $\vy_{:t}$.  Analogously, our $G_t+H_t$ is the log of the total exponentiated scores of \emph{all} sequences with prefix $\vy_{:t}$.  $G_t$ and $H_t$
  might be called the \emph{logprob-so-far} and \emph{logprob-to-go} of $\vy_{:t}$.

Just as $A^*$ approximates $h$ with a ``heuristic'' $\hat{h}$,
the next section will approximate $H_t$ using a neural estimate $\hat{H}_t$ (\crefrange{eq:Hhat}{eq:sbar}).
However, the specific form of our approximation is inspired by cases where $H_t$ can be computed exactly.  We consider those in the remainder of this section.

\subsection{Exact sampling from HMMs}
\label{sec:hmm}

A \emph{hidden Markov model} (HMM) specifies a normalized \emph{joint} distribution $p_\theta(\vx,\vy) = \exp G_T$ over state sequence $\vy$ and observation sequence $\vx$,\footnote{The HMM actually specifies a distribution over a pair of infinite sequences, but here we consider the marginal distribution over just the length-$T$ prefixes.}  Thus the posterior $p_\theta(\vy \mid \vx)$ is proportional to $\exp G_T$,
as required by \cref{eq:condit}.

The HMM specifically defines $G_T$ by \crefrange{eq:G}{eq:s} with $\vs_t = y_t$ and $g_\theta(\vs_{t-1},x_t,y_t) = \log p_\theta(y_t \mid y_{t-1}) + \log p_\theta(x_t \mid y_t)$.\footnote{\label{fn:bos}It takes $\vs_0 = \bos$, a beginning-of-sequence symbol, so $p_\theta(y_1 \mid \bos)$ specifies the initial state distribution.}

In this setting, $H_t$ can be computed exactly by the \emph{backward
  algorithm} \cite{rabiner-1989}.
(Details are given in \cref{app:hmm-backward} for completeness.)

\subsection{Exact sampling from OOHMMs}\label{sec:oohmm}

For sequence tagging, a weakness of (first-order) HMMs is that the model state $\vs_t = y_t$ may contain little information: only the most recent tag $y_t$ is remembered, so the number of possible model states $\vs_t$ is limited by the vocabulary of output tags.

We may generalize so that the data generating process is in a latent state $u_t \in \{1,\ldots,k\}$ at each time $t$, and the observed $y_t$---along with $x_t$---is generated from $u_t$.  Now $k$ may be arbitrarily large.  The model has the form
\begin{align}
p_\theta(\vx,\vy) &= \exp G_T   \label{eq:oohmm} \\
  & = \sum_{\vu} \prod_{t=1}^T p_\theta(u_t \mid u_{t-1}) \cdot p_\theta(x_t,y_t \mid u_t) \nonumber
\end{align}
This is essentially a pair HMM \cite{knudsen2003} without insertions or deletions, also known as an ``$\epsilon$-free'' or ``same-length'' probabilistic finite-state transducer.  We refer to it here as an \emph{output-output HMM} (OOHMM).\footnote{\label{fn:iohmm}This is by analogy with the \emph{input-output HMM} (IOHMM) of \newcite{bengio-frasconi-1996}, which defines $p(\vy \mid \vx)$ directly and conditions the transition to $u_t$ on $x_t$.  The OOHMM instead defines $p(\vy \mid \vx)$ by conditionalizing \eqref{eq:oohmm}---which avoids the \emph{label bias} problem \cite{lafferty-mccallum-pereira-2001} that in the IOHMM, $y_t$ is independent of future input $\vx_{t:}$ (given the past input $\vx_{:t}$).}

Is this still an example of the general model architecture from \cref{sec:setup}?
Yes.  Since $u_t$ is latent and evolves stochastically, it cannot be used as the state $\vs_t$ in \crefrange{eq:G}{eq:s} or \eqref{eq:noindep}.  However, we {\em can} define $\vs_t$ to be
the model's \emph{belief state} after observing $(\vx_{:t},\vy_{:t})$.  The belief state is the posterior probability distribution over the
underlying state $u_t$ of the system.  That is, $\vs_t$ deterministically keeps track of all possible states that the OOHMM might be in---just as the state of a determinized FSA keeps track of all possible states that the original nondeterministic FSA might be in.

We may compute the belief state in terms of a vector of \emph{forward probabilities} that starts at $\valpha_0$,
\begin{align}
(\valpha_0)_u &\defeq \begin{cases}
  1 & \text{if }u=\bos\text{ (see \cref{fn:bos})} \\
  0 & \text{if }u=\text{any other state}
\end{cases}
\raisetag{16pt}
\end{align}
and is updated deterministically for each $0 < t \leq T$ by the \emph{forward algorithm} \cite{rabiner-1989}:
\begin{align}
(\valpha_t)_u &\defeq \sum_{u'=1}^k (\valpha_{t-1})_{u'} \cdot p_\theta(u \mid u') \cdot p_\theta(x_t,y_t \mid u) \raisetag{10pt} \label{eq:forward}
\end{align}

\vspace{-3pt}\noindent
$(\valpha_t)_u$ can be interpreted as the logprob-so-far \emph{if} the system is in state $u$ after observing $(\vx_{:t},\vy_{:t})$.  We may express the update rule \eqref{eq:forward} by $\valpha_t^\top = \valpha_{t-1}^\top P$ where
the matrix $P$ depends on $(x_t,y_t)$, namely $P_{u'u} \defeq p_\theta(u \mid u') \cdot p_\theta(x_t,y_t \mid u)$.

The belief state $\vs_t \defeq \norm{\valpha_t} \in \Real^k$ simply normalizes $\valpha_t$ into a probability vector, where $\norm{\vecc{u}} \defeq \vecc{u}/(\vecc{u}^\top \vone)$ denotes the \emph{normalization operator}.
The state update \eqref{eq:forward} now takes the form \eqref{eq:s} as desired, with $f_\theta$ a normalized vector-matrix product:
\begin{align}
  \vs_t^\top &= f_\theta(\vs_{t-1},x_t,y_t) \defeq \norm{\vs_{t-1}^\top P} \label{eq:s-update}
\end{align}

As in the HMM case, we define $G_t$ as the log of the generative prefix probability,
\begin{align}
G_t &\defeq \log p_\theta(\vx_{:t},\vy_{:t})
= \log \textstyle \sum_u (\valpha_t)_u \label{eq:sum-alpha}
\end{align}
which has the form \eqref{eq:G} as desired if we put
\begin{align}
g_\theta(\vs_{t-1},x_t,y_t) &\defeq G_t - G_{t-1} \label{eq:g} \\
&= \log \frac{\valpha_{t-1}^\top P \vone}{\valpha_{t-1}^\top\vone}
= \log \; (\vs_{t-1}^\top P \vone) \nonumber
\end{align}

Again, exact sampling is possible.  It suffices to compute \eqref{eq:GplusH}.
For the OOHMM, this is given by
\begin{equation}
\textstyle \sum_{\vy_{t:}} \exp G_T = \valpha_t^\top \vbeta_t \label{eq:alpha-beta}
\end{equation}
where $\vbeta_T\defeq\vone$
and the \emph{backward algorithm}
\begin{align}
(\vbeta_t)_v \label{eq:beta-oohmm}
&\defeq p_\theta(\vx_{t:} \mid u_t=u) \\
&=\sum_{\vu_{t:},\vy_{t:}} p_\theta(\vu_{t:},\vx_{t:},\vy_{t:} \mid u_t=u) \nonumber \\
&= \sum_{u'} \underbrace{p_\theta(u' \mid u) \cdot p(x_{t+1} \mid
  u')}_{\text{call this }\Pbar_{uu'}} \cdot (\vbeta_{t+1})_{u'} \nonumber
\end{align}
for $0 \leq t < T$ uses dynamic programming to find the total probability of all ways to generate the future observations $\vx_{t:}$.  Note that $\valpha_t$ is defined for a \emph{specific prefix} $\vy_{:t}$ (though it sums over all $\vu_{:t}$), whereas
$\vbeta_t$ sums over \emph{all suffixes} $\vy_{t:}$ (and over all $\vu_{t:}$), to achieve the asymmetric summation in \eqref{eq:alpha-beta}.

Define $\vsbar_t \defeq \norm{\vbeta_t} \in \Real^k$ to be a
normalized version of $\vbeta_t$.  The $\vbeta_t$ recurrence
\eqref{eq:beta-oohmm} can clearly be expressed in the form $\vsbar_t =
\norm{\Pbar\vsbar_{t+1}}$, much like \eqref{eq:s-update}.\looseness=-1

\subsection{The logprob-to-go for OOHMMs}

Let us now work out the definition of $H_t$ for OOHMMs
(cf.~\cref{eq:H-as-diff} in \cref{app:hmm-backward} for HMMs).  We will write it in terms of $\hat{H}_t$ from \cref{sec:Hhat}.
Let us define $\hat{H}_t$ symmetrically to $G_t$ (see \eqref{eq:sum-alpha}):
\begin{align}
\hat{H}_t &\defeq \log \sum_u (\vbeta_t)_u \;\;(= \log \vone^\top \vbeta_t) \label{eq:sum-beta}
\end{align}
which has the form \eqref{eq:Hhat} as desired if we put
\begin{align}
h_\phi(\vsbar_{t+1},x_{t+1}) &\defeq \hat{H}_t - \hat{H}_{t+1} = \log\;( \vone^\top \Pbar \vsbar_{t+1}) \label{eq:h}
\end{align}
From \cref{eq:H,eq:alpha-beta,eq:sum-alpha,eq:sum-beta}, we see
\begin{align}
H_t &= \log \big( \sum_{\vy_{t:}} \exp G_T \big) - G_t \nonumber \\
    &= \log \frac{\valpha_t^\top \vbeta_t}{(\valpha_t^\top \vone)(\vone^\top \vbeta_t)} + \log\;(\vone^\top \vbeta_t) \nonumber \\
    &= \underbrace{\log \vs_t^\top \vsbar_t}_{\text{call this $C_t$}} + \hat{H}_t \label{eq:C}
\end{align}
where $C_t \in \Real$ can be regarded as evaluating the \emph{compatibility} of the state distributions $\vs_t$ and $\vsbar_t$.

In short, the generic strategy \eqref{eq:q-exact-unnorm} for exact sampling says that for an OOHMM, $y_t$ is distributed as\looseness=-1
\begin{align}
\lefteqn{p_\theta(y_t \mid \vx, \vy_{:t-1})
 \propto \exp\; (g_\theta(\vs_{t-1},x_t,y_t) + H_t)} \nonumber \\
 &\propto
\exp\; (\underbrace{g_\theta(\vs_{t-1},x_t,y_t)}_{\ \ \text{depends on }\vx_{:t},\vy_{:t}\ \ }
   + \underbrace{C_t}_{\text{on }\vx,\vy_{:t}}
   + \underbrace{\hat{H}_t}_{\text{on }\vx_{t:}}
) \nonumber \\
&\propto \exp\; (g_\theta(\vs_{t-1},x_t,y_t) + C_t) \label{eq:q-exact-oohmm}
\end{align}
This is equivalent to choosing $y_t$ in proportion to \eqref{eq:alpha-beta}---but we now turn to settings where it is infeasible to compute \eqref{eq:alpha-beta} exactly.  There we will use the formulation \eqref{eq:q-exact-oohmm} but approximate $C_t$.  For completeness, we will also consider how to approximate $\hat{H}_t$, which dropped out of the above distribution (because it was the same for all choices of $y_t$) but may be useful for other algorithms (see \cref{sec:sis}).

\section{Neural Modeling as Approximation}\label{sec:approx}

\subsection{Models with large state spaces}

The expressivity of an OOHMM is limited by the number of states $k$.  The state $u_t \in \{1,\ldots,k\}$ is a bottleneck between the past $(\vx_{:t},\vy_{:t})$ and the future $(\vx_{t:},\vy_{t:})$, in that past and future are \emph{conditionally independent} given $u_t$.  Thus, the mutual information between past and future is  at most $\log_2 k$ bits.\cutforspace{\footnote{Another perspective is that if all $p(\text{future} \mid \text{past})$ values were laid out in an infinite matrix whose row labels were possible pasts $(\vx_{:t},\vy_{:t})$ and whose column labels were possible futures $(\vx_{t:},\vy_{t:})$, then---strikingly---this matrix would have low rank $\leq k$.  Why?  Any row must be a linear combination of the same $k$ shared distributions $\{p(\text{future} \mid v): 1 \leq u \leq k\}$. (Specifically, for the row $(\vx_{:t},\vy_{:t})$, the $v$\textsuperscript{th} shared distribution gets coefficient $p(u_t=u \mid \text{past}=(\vx_{:t},\vy_{:t}))$, that is, $(\vs_t)_u$.)}}

In many NLP domains, however, the past seems to carry substantial information about the future.  The first half of a sentence greatly reduces the uncertainly about the second half, by providing information about topics, referents, syntax, semantics, and discourse.  This suggests that an accurate HMM language model $p(\vx)$ would require \emph{very large $k$}---as would a generative OOHMM model $p(\vx,\vy)$ of \emph{annotated} language.  The situation is perhaps better for discriminative models $p(\vy\mid\vx)$, since much of the information for predicting $\vy_{t:}$ might be available in $\vx_{t:}$.  Still, it is important to let $(\vx_{:t},\vy_{:t})$ contribute enough additional information about $\vy_{t:}$: even for short strings, making $k$ too small (giving $\leq \log_2 k$ bits) may harm prediction \cite{dreyer-smith-eisner-2008}.

Of course, \eqref{eq:noindep} says that an OOHMM can express any joint distribution for which the mutual information is finite,\footnote{This is not true for the language of balanced parentheses.} by taking $k$ large enough for $v_{t-1}$ to capture the relevant info\cutforspace{rmation} from $(\vx_{:t-1},\vy_{:t-1})$.

So why not just take $k$ to be large---say, $k=2^{30}$ to allow 30 bits of information?  Unfortunately, evaluating $G_T$ then becomes very expensive---both computationally and statistically.  As we have seen, if we define $\vs_t$ to be the belief state $\norm{\valpha_t} \in \Real^k$, updating it at each observation $(x_t,y_t)$ (\cref{eq:s}) requires multiplication by a $k \times k$ matrix $P$.  This takes time $O(k^2)$, and requires enough data to learn $O(k^2)$ transition probabilities.

\subsection{Neural approximation of the model}
\label{sec:neural-model}
\label{sec:globally}

As a solution, we might hope that for the inputs $\vx$ observed in practice, the very high-dimensional belief states $\norm{\valpha_t} \in \Real^k$ might tend to lie near a $d$-dimensional manifold where $d \ll k$.  Then we could take $\vs_t$ to be a vector in $\Real^d$ that compactly encodes the approximate coordinates of $\norm{\valpha_t}$ relative to the manifold: $\vs_t = \nu(\norm{\valpha_t})$, where $\nu$ is the encoder.

In this new, nonlinearly warped coordinate system, the functions
of $\vs_{t-1}$
in \eqref{eq:G}--\eqref{eq:s} are no longer the simple, essentially linear functions
given by \eqref{eq:s-update} and \eqref{eq:g}.
They become nonlinear functions operating on the manifold coordinates. ($f_\theta$
in \eqref{eq:s-update}
should now ensure that $\vs_t^\top \approx \nu(\norm{(\nu^{-1}(\vs_{t-1}))^\top P})$, and $g_\theta$
in \eqref{eq:g}
should now estimate $\log\; (\nu^{-1}(\vs_{t-1}))^\top P \vone$.)  In
a sense, this is the reverse of the ``kernel trick'' \citep{boser1992} that converts a low-dimensional nonlinear function to a high-dimensional linear one.

Our hope is that $\vs_t$ has enough dimensions $d \ll k$ to capture the useful information from the true $\norm{\valpha_t}$, \textbf{and} that $\theta$ has enough dimensions $\ll k^2$ to capture most of the dynamics of \cref{eq:s-update,eq:g}.  We thus proceed to fit the neural networks $f_\theta, g_\theta$ directly to the data, {\em without ever knowing} the true $k$ or the structure of the original operators $P \in \Real^{k\times k}$.

We regard this as the implicit justification for various published probabilistic sequence models $p_\theta(\vy \mid \vx)$ that incorporate neural networks.  These models usually have the form of \cref{sec:setup}.   Most simply, $(f_\theta,g_\theta)$ can be instantiated as one time step in an RNN \cite{aharoni2016},  but it is common to use enriched versions such as deep LSTMs.  It is also common to have the state $\vs_t$ contain not only a vector of manifold coordinates in $\Real^d$ but also some unboundedly large representation of $(\vx,\vy_{:t})$ (cf.\@ \cref{eq:noindep}), so the $f_\theta$ neural network can refer to this material with an attentional \cite{bahdanau2014} or stack mechanism \cite{dyer2015}.

A few such papers have used \emph{globally} normalized conditional models that can be viewed as approximating some OOHMM, e.g., the parsers of \newcite{dyer2016} and \newcite{andor-et-al-2016-acl}.  That is the case (\cref{sec:setup}) that particle smoothing aims to support.  Most papers are \emph{locally} normalized conditional models \cite[e.g.,][]{kann2016,aharoni2016}; these simplify supervised training and can be viewed as approximating IOHMMs (\cref{fn:iohmm}).  For locally normalized models, $H_t=0$ by construction, in which case particle filtering (which estimates $H_t=0$) is just as good as particle smoothing.
Particle filtering is still useful for these models, but lookahead's inability to help them is an expressive limitation (known as \emph{label bias}) of locally normalized models.  We hope the existence of particle smoothing (which learns an estimate $H_t$) will make it easier to adopt, train, and decode globally normalized models, as discussed in \cref{sec:apps}.
\looseness=-1

\subsection{Neural approximation of logprob-to-go}\label{sec:q}

We can adopt the same neuralization trick to approximate the OOHMM's logprob-to-go $H_t = C_t + \hat{H}_t$.  We take $\vsbar_t \in \Real^d$ on the same theory that it is a low-dimensional
reparameterization of $\norm{\beta_t}$, and define $(\bar{f}_\phi,h_\phi)$ in \crefrange{eq:Hhat}{eq:sbar} to be neural networks.  Finally, we must replace the definition of $C_t$ in \eqref{eq:C} with another neural network $c_\phi$ that works on the low-dimensional approximations:\footnote{\label{fn:CT}$C_T=0$ is correct according to \eqref{eq:C}.  Forcing this ensures $H_T=0$, so our approximation becomes exact as of $t=T$.}
\begin{align}
  C_t &\defeq c_\phi(\vs_t,\vsbar_t) &\text{(except that $C_T\defeq 0$)}\label{eq:c}
\end{align}
The resulting approximation to \eqref{eq:q-exact-oohmm} (which does not actually
require $h_\phi$) will be denoted $q_{\theta,\phi}$:
\begin{align}\label{eq:q-local}
q_{\theta,\phi}(y_t \mid \vx, \vy_{:t-1}) \defpropto \exp\; (g_\theta(\vs_{t-1},x_t,y_t) + C_t)
\end{align}

The neural networks in the present section are all parameterized by $\phi$, and are intended to produce an estimate of the logprob-\emph{to-go} $H_t$---a function of $\vx_{t:}$, which sums over all possible $\vy_{t:}$.

By contrast, the OOHMM-inspired neural networks suggested in \cref{sec:neural-model} were used to specify an actual model of the logprob-\emph{so-far} $G_t$---a function of $\vx_{:t}$ and $\vy_{:t}$---using separate parameters $\theta$.\looseness=-1

Arguably $\phi$ has a harder modeling job than $\theta$ because it must implicitly sum over possible futures $\vy_{t:}$.  We now consider how to get corrected samples from $q_{\theta,\phi}$ even if $\phi$ gives poor estimates of $H_t$, and then how to train $\phi$ to improve those estimates.

\section{Particle smoothing}
\label{sec:sis}

In this paper, we assume nothing about the given model $G_T$ except that it is given in the form of \crefrange{eq:condit}{eq:s} (including the parameter vector $\theta$).\looseness=-1

Suppose we run the exact sampling strategy but approximate $p_\theta$ in \eqref{eq:q-exact} with a \emph{proposal distribution} $q_{\theta,\phi}$ of the form in \eqref{eq:c}--\eqref{eq:q-local}.
Suppressing the subscripts on $p$ and $q$ for brevity, this means we are effectively drawing $\vy$ not from $p(\vy \mid \vx)$ but from
\begin{align}\label{eq:q}
q(\vy \mid \vx) &= \prod_{t=1}^T q(y_t \mid \vx, \vy_{:t-1})
\end{align}
If $C_t \approx H_t + \text{const}$ within each $y_t$ draw,
then $q \approx p$.

\emph{Normalized importance sampling} corrects (mostly) for the approximation by drawing \emph{many} sequences $\vy\psup{1}, \ldots \vy\psup{M}$ IID from \eqref{eq:q} and assigning $\vy\psup{m}$ a relative \emph{weight} of $w\psup{m} \defeq \frac{p(\vy\psup{m} \mid \vx)}{q(\vy\psup{m} \mid \vx)}$.  This \emph{ensemble of weighted particles} yields a distribution
\begin{align}
\phat(\vy) &\defeq \tfrac{\sum_{m=1}^{M} w\psup{m} \mathbb{I}(\vy = \vy\psup{m})}{\sum_{m=1}^{M} w\psup{m}} \approx p(\vy \mid \vx)
\label{eq:phat}
\end{align}
that can be used as discussed in \cref{sec:apps}.  To compute $w\psup{m}$ in practice, we replace the numerator $p(\vy\psup{m} \mid \vx)$ by the unnormalized version $\exp G_T$, which gives the same $\phat$.  Recall that each $G_T$ is a sum $\sum_{t=1}^T g_\theta(\cdots)$.

\emph{Sequential importance sampling} is an equivalent implementation that makes $t$ the \emph{outer} loop and $m$ the \emph{inner} loop.  It computes a \emph{prefix ensemble}
\begin{align}
Y_t &\defeq \{(\vy_{:t}\psup{1},w_t\psup{1}),\ldots,(\vy_{:t}\psup{M},w_t\psup{M})\}
\end{align}
for each $0 \leq t \leq T$ in sequence.
Initially, $(\vy_{:0}\psup{m},w_0\psup{m}) = (\epsilon,\exp C_0)$ for all $m$.
Then for $0 < t \leq T$, we extend these particles in
parallel:
\begin{align}
\vy_{:t}\psup{m} &= \vy_{:t-1}\psup{m} y_t\psup{m} \hspace{13mm}\text{(concatenation)}\\
w_t\psup{m} &= w_{t-1}\psup{m} \; \tfrac{\exp\; (g_\theta(\vs_{t-1},x_t,y_t)\,+\,C_t\,-\,C_{t-1})}{q(y_t \mid \vx, \vy_{:t-1})} \label{eq:sis}
\end{align}
where each $y_t\psup{m}$ is drawn from \eqref{eq:q-local}.
Each $Y_t$ yields a distribution $\phat_t$ over prefixes $\vy_{:t}$, which estimates the distribution $p_t(\vy_{:t}) \defpropto \exp\;(G_t+C_t)$.
We return $\phat \defeq \phat_T \approx p_T = p$.  This
gives the same $\phat$ as in \eqref{eq:phat}: the final
$\vy_T\psup{m}$ are the same, with the same final weights
$w_T\psup{m} = \frac{\exp G_T}{q(\vy\psup{m} \mid \vx)}$,
where $G_T$ was now summed up
as $C_0 + \sum_{t=1}^T g_\theta(\cdots) + C_t - C_{t-1}$.

That is our basic \emph{particle smoothing} strategy.  If we use the naive approximation $C_t=0$ everywhere, it reduces to \emph{particle filtering}.
In either case, various well-studied
improvements become available, such as various resampling schemes \cite{douc2005} and the particle cascade \cite{paige2014}.\footnote{The particle cascade would benefit from an estimate of $\hat{H}_t$, as it (like A$^*$ search) compares particles of different lengths.}

An easy improvement is \emph{multinomial resampling}.  After computing each $\phat_t$, this replaces $Y_t$ with a set of $M$ new draws from $\phat_t$ $(\approx p_t)$, each of weight 1---which tends to drop low-weight particles and duplicate high-weight ones.\footnote{While resampling mitigates the degeneracy problem, it could also reduce the diversity of particles. In our experiments in this paper, we only do multinomial resampling when the effective sample size of $\hat{p}_t$ is lower than $\frac{M}{2}$. \citet{doucet2009} give a more thorough discussion on when to resample.}  For this to usefully focus the ensemble on good prefixes $\vy_{:t}$, $p_t$ should be a good approximation to the true marginal $p(\vy_{:t} \mid \vx) \propto \exp\; (G_t+H_t)$ from \eqref{eq:H}.  That is why we arranged for $p_t(\vy_{:t}) \propto  \exp\;(G_t+C_t)$.
 Without $C_t$, we would have only $p_t(\vy_{:t}) \propto \exp G_t$---which is fine for the traditional particle filtering setting, but in our setting it ignores future information in $\vx_{t:}$ (which we have assumed is available) and also favors sequences $\vy$ that happen to accumulate most of their global score $G_T$ early rather than late (which is possible when the globally normalized model \eqref{eq:condit}--\eqref{eq:G} is \emph{not} factored in the generative form \eqref{eq:noindep}).
\looseness=-1

\section{Training the Sampler Heuristic}
\label{sec:optimization}
We now consider training the parameters $\phi$ of our sampler.  These
parameters determine the updates $\bar{f}_\phi$ in \eqref{eq:sbar} and the compatibility function $c_\phi$ in \eqref{eq:c}.  As a result, they determine the proposal
distribution $q$ used in \cref{eq:q,eq:sis},
and thus determine the stochastic choice of $\phat$ that is returned by
the sampler on a given input $\vx$.

In this paper, we simply try to tune $\phi$ to yield good proposals.  Specifically, we try to ensure that $q_\phi(\vy \mid \vx)$ in \cref{eq:q} is close to $p(\vy \mid \vx)$ from \cref{eq:condit}.
While this may not be necessary for the sampler to perform well downstream,\footnote{\label{fn:end2end}In principle, one could attempt to train $\phi$ ``end-to-end'' on some downstream objective by using reinforcement learning or the Gumbel-softmax trick \cite{jang2016,maddison2016}.  For example, we might try to ensure that $\phat$ closely matches the model's distribution $p$ (\cref{eq:phat})---the ``natural'' goal of sampling.  This objective can tolerate inaccurate local proposal distributions in cases where the algorithm could recover from them through resampling.  Looking even farther downstream, we might merely want $\phat$---which is typically used to compute expectations---to provide accurate guidance to some decision or training process (see \cref{app:apps}).  This might not require fully matching the model, and might even make it desirable to deviate from an inaccurate model.} it does guarantee it (assuming that the model $p$ is correct).
Specifically, we seek to minimize
\begin{equation}
(1 - \lambda) \text{KL}(p || q_\phi) + \lambda \text{KL}(q_\phi || p)\ \ \ \text{(with $\lambda \in [0,1]$)}
\end{equation}
averaged over examples $\vx$ drawn from a training set.\footnote{Training a single approximation $q_\phi$ for all $\vx$ is known as {\em amortized inference}.}
(The training set need not provide true $\vy$'s.)

The {\em inclusive KL divergence} $\text{KL} \left( p || q_{\phi} \right)$ is an expectation under $p$.  We estimate it by replacing $p$ with a sample $\phat$, which in practice we can obtain with our sampler under the current $\phi$.  (The danger, then, is that $\phat$ will be biased when $\phi$ is not yet well-trained; this can be mitigated by increasing the sample size $M$ when drawing $\phat$ for training purposes.)

Intuitively, this term tries to encourage $q_\phi$ in future to re-propose those $\vy$ values that turned out to be ``good'' and survived into $\phat$ with high weights.

The {\em exclusive KL divergence} $\text{KL}(q_\phi || p)$ is an expectation under $q_\phi$.
Since we can sample from $q_\phi$ exactly, we can get an unbiased estimate of $\nabla_\phi \text{KL}(q_\phi || p)$
with the likelihood ratio trick \citep{glynn1990}.\footnote{The normalizing constant of $p$ from \eqref{eq:condit} can be ignored because the gradient of a constant is 0.}
(The danger is that such ``REINFORCE'' methods tend to suffer from very
high variance.)

This term is a popular objective for variational approximation.  Here, it tries to discourage $q_\phi$ from re-proposing ``bad'' $\vy$ values that turned out to have low $\exp G_T$ relative to their proposal probability.

Our experiments balance ``recall'' (inclusive) and ``precision'' (exclusive) by taking $\lambda = \frac{1}{2}$ (which \Cref{app:lastchar} compares to $\lambda \in \{0,1\})$.
Alas, because of our approximation to the inclusive term, neither term's gradient will ``find'' and directly encourage good $\vy$ values that have never been proposed. \Cref{app:gradient-derivation} gives further discussion and formulas.

\section{Models for the Experiments}
\label{sec:model}

To evaluate our methods, we needed pre-trained models $p_\theta$.  We experimented
on several models.  In each case, we trained a {\em generative} model $p_\theta(\vx,\vy)$,
so that we could try sampling from its posterior distribution $p_\theta(\vy \mid \vx)$.  This is a very common setting where particle smoothing
should be able to help.  Details for replication are given in \cref{app:model-details}.

\subsection{Tagging models}
\label{sec:tagging}

We can regard a tagged sentence $(\vx,\vy)$ as a string over the
``pair alphabet'' $\mathcal{X} \times \mathcal{Y}$.  We train an RNN
language model over this ``pair alphabet''---this is a neuralized OOHMM as suggested in \cref{sec:neural-model}:
\begin{align}
\log p_\theta(\vx, \vy) &= \sum_{t=1}^{T} \log p_\theta(x_t,y_t \mid \vs_{t-1})
\end{align}

This model is locally normalized, so that $\log p_\theta(\vx, \vy)$ (as well as its gradient) is straightforward to compute for a given training pair $(\vx, \vy)$.  Joint sampling from it would also be easy (\cref{sec:globally}).

However, $p(\vy \mid \vx)$ is globally renormalized (by an unknown partition function that depends on $\vx$, namely $\exp H_0$).  Conditional sampling of $\vy$ is therefore potentially hard.  Choosing $y_t$ optimally requires knowledge of $H_t$, which depends on the future $\vx_{t:}$.

As we noted in \cref{sec:intro}, many NLP tasks can be seen as tagging problems. In this paper we experiment with two such tasks: \textbf{English stressed syllable tagging}, where the stress of a syllable often depends on the number of remaining syllables,\footnote{English, like many other languages, assigns stress from right to left \cite{hayes1995}.}
providing good reason to use the \emph{lookahead} provided by particle smoothing; and \textbf{Chinese NER}, which is a familiar textbook application and reminds the reader that our formal setup (tagging) provides enough machinery to treat other tasks (chunking).

\paragraph{English stressed syllable tagging}
This task tags a sequence of phonemes $\vx$, which form a word, with their stress markings $\vy$.
Our training examples are the stressed words in the CMU pronunciation dictionary \cite{cmudict}.
We test the sampler on held-out unstressed words.

\paragraph{Chinese social media NER}
This task does named entity recognition in Chinese, by tagging the characters of a Chinese sentence in a way that marks the named entities.  We use the dataset from \citet{peng2015}, whose tagging scheme is a variant of the BIO scheme mentioned in \cref{sec:intro}.
  We test the sampler on held-out sentences.

\subsection{String source separation}
\label{sec:sourcesep}

This is an artificial task that provides a discrete analogue of speech source separation \cite{zibulevsky2001}.
The generative model is that $J$ strings (possibly of different lengths) are generated IID from an RNN language model, and are then combined into a single string $\vx$ according to a random {\em interleaving} string $\vy$.\footnote{We formally describe the generative process in \cref{app:sourcesep-generation}.} The posterior $p(\vy \mid \vx)$ predicts the interleaving string, which suffices to reconstruct the original strings.
The interleaving string is selected from the uniform distribution over all possible interleavings (given the $J$ strings' lengths).
For example, with $J=2$, a possible generative story is that we first sample two strings \textcolor{red}{Foo} and \textcolor{blue}{Bar} from an RNN language model. We then draw an interleaving string \textcolor{red}{1}\textcolor{red}{1}\textcolor{blue}{2}\textcolor{red}{1}\textcolor{blue}{2}\textcolor{blue}{2} from the aforementioned uniform distribution, and interleave the $J$ strings deterministically to get \textcolor{red}{F}\textcolor{red}{o}\textcolor{blue}{B}\textcolor{red}{o}\textcolor{blue}{a}\textcolor{blue}{r}.

$p(\vx,\vy)$ is proportional to the product of the probabilities of the $J$ strings.  The only parameters of $p_\theta$, then, are the parameters of the RNN language model, which we train on clean (non-interleaved) samples from a corpus.  We test the sampler on random interleavings of held-out samples.

The state $\vs$ (which is provided as an input to $c_\theta$ in \eqref{eq:c}) is the concatenation of the $J$ states of the language model as it independently generates the $J$ strings, and $g_\theta(\vs_{t-1},x_t,y_t)$ is the log-probability of generating $x_t$ as the next character of the $y_t$\textsuperscript{th} string, given that string's language model state within $\vs_{t-1}$.  As a special case, $\vx_T=\eos$ (see \cref{fn:eos}), and $g_\theta(\vs_{T-1},\eos,\eos)$ is the total log-probability of termination in all $J$ language model states.

String source separation has good reason for lookahead: appending character ``o'' to a reconstructed string ``\textvisiblespace gh'' is only advisable if ``s'' and ``t'' are coming up soon to make ``ghost.'' It also illustrates a powerful application setting---posterior inference under a generative model.  This task conveniently allowed us to construct the generative model from a pre-trained language model.  Our constructed generative model illustrates that the state $\vs$ and transition function $f$ can reflect interesting problem-specific structure. 

\paragraph{CMU Pronunciation dictionary}
The CMU pronunciation dictionary (already used above) provides sequences of phonemes. Here we use words no longer than $5$ phonemes. We interleave the (unstressed) phonemes of $J=5$ words.

\paragraph{Penn Treebank}
The PTB corpus \cite{marcus1993} provides English sentences, from which we use only the sentences of length $\leq 8$.
  We interleave the words of $J=2$ sentences.

\begin{figure*}
\begin{subfigure}[t]{.25\textwidth}
\includegraphics[width=0.97\textwidth]{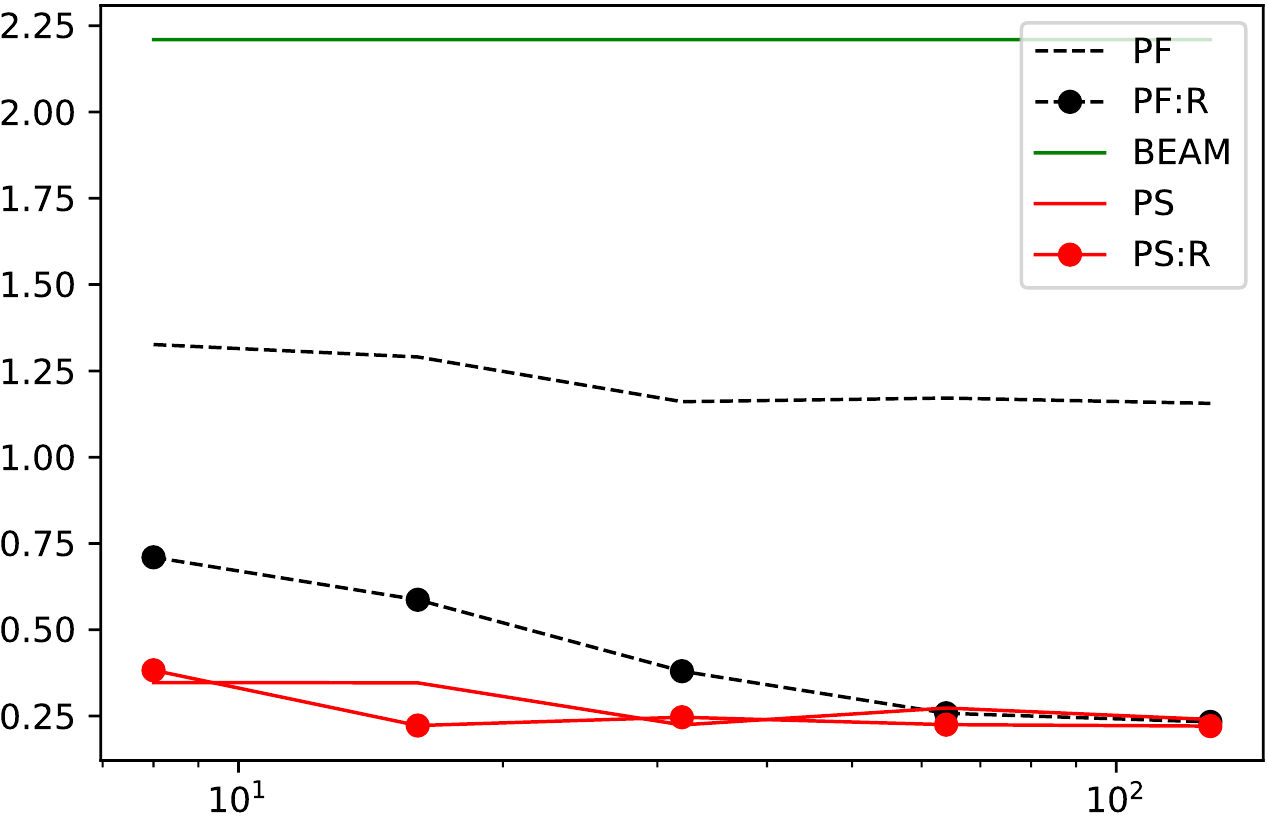}
\caption{tagging: stressed syllables}
\label{fig:divcmu}
\end{subfigure} \hspace{-1em}
\ \
\begin{subfigure}[t]{.25\textwidth}
\includegraphics[width=0.97\textwidth]{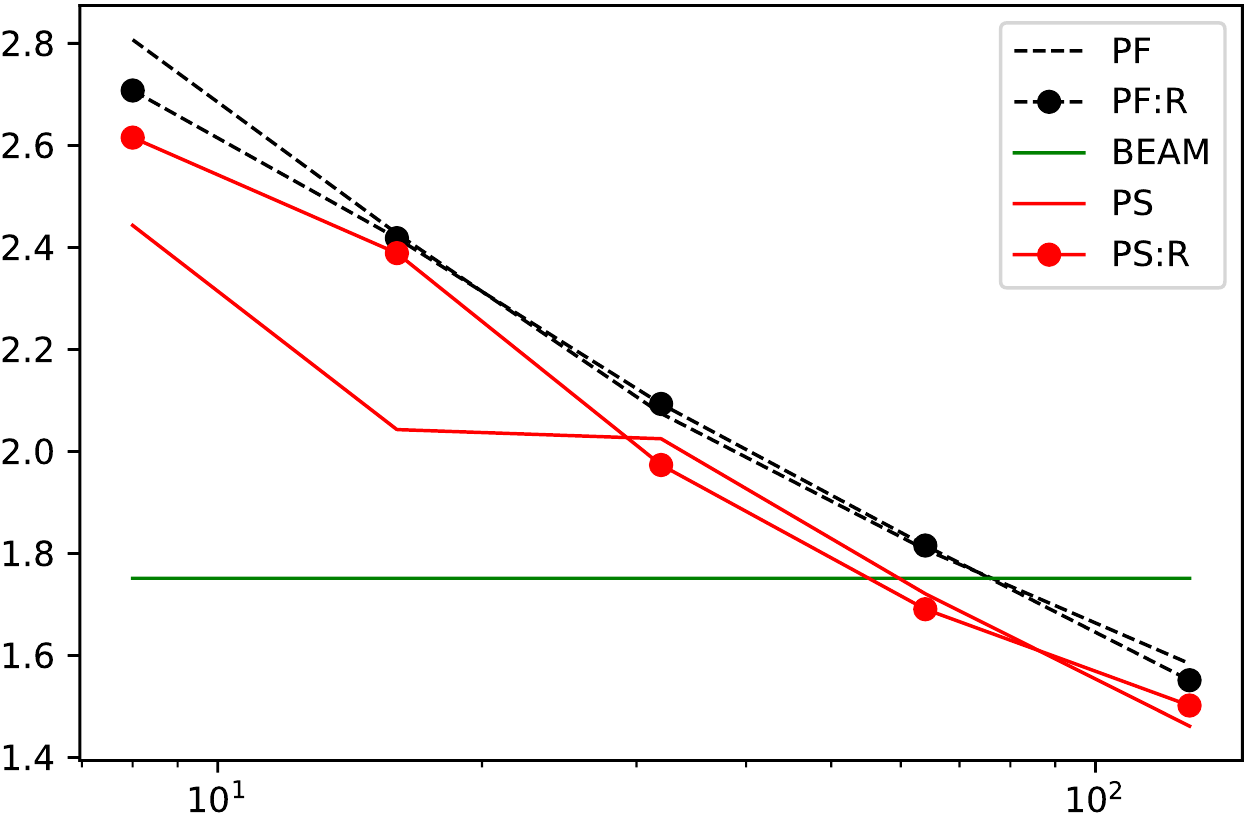}
\caption{tagging: Chinese NER}
\label{fig:divweibo}
\end{subfigure} \hspace{-1em}
\ \
\begin{subfigure}[t]{.25\textwidth}
\includegraphics[width=0.97\textwidth]{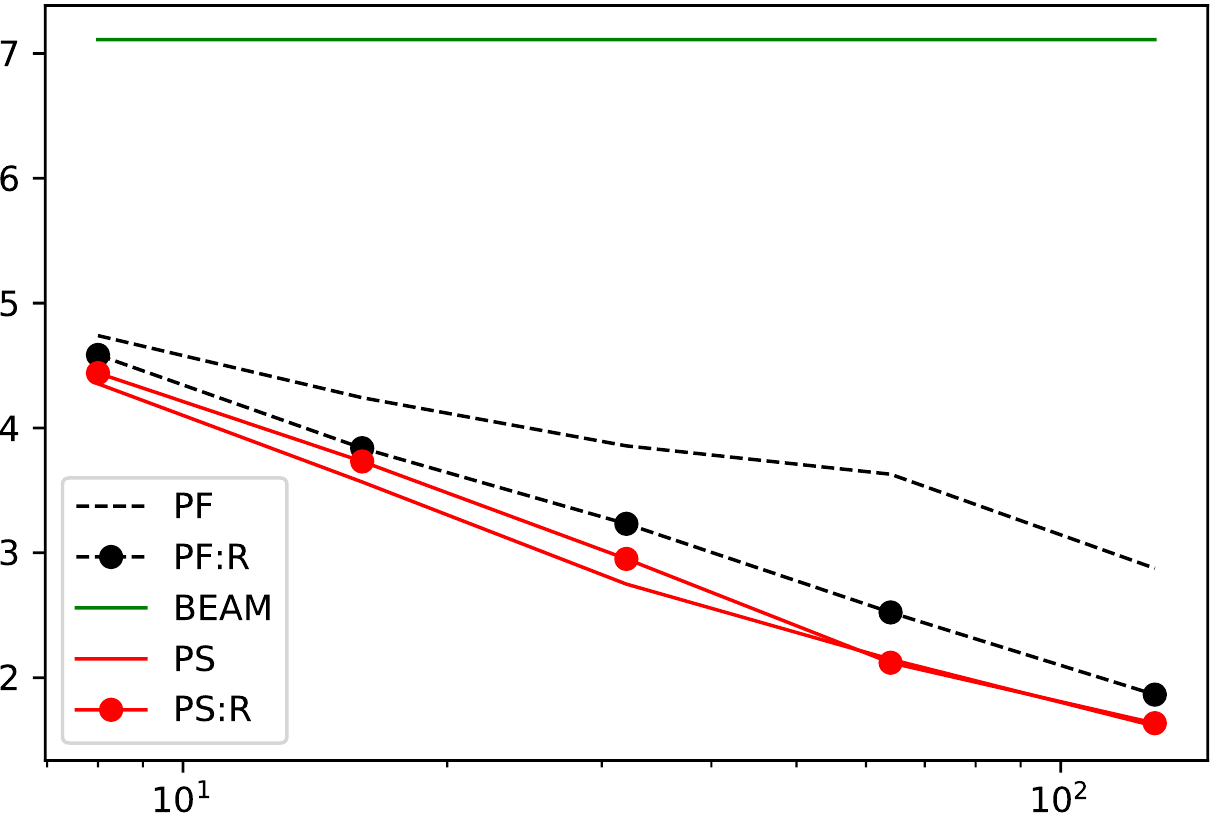}
\caption{source separation: PTB}
\label{fig:div}
\end{subfigure} \hspace{-1em}
\ \
\begin{subfigure}[t]{.25\textwidth}
\includegraphics[width=0.97\textwidth]{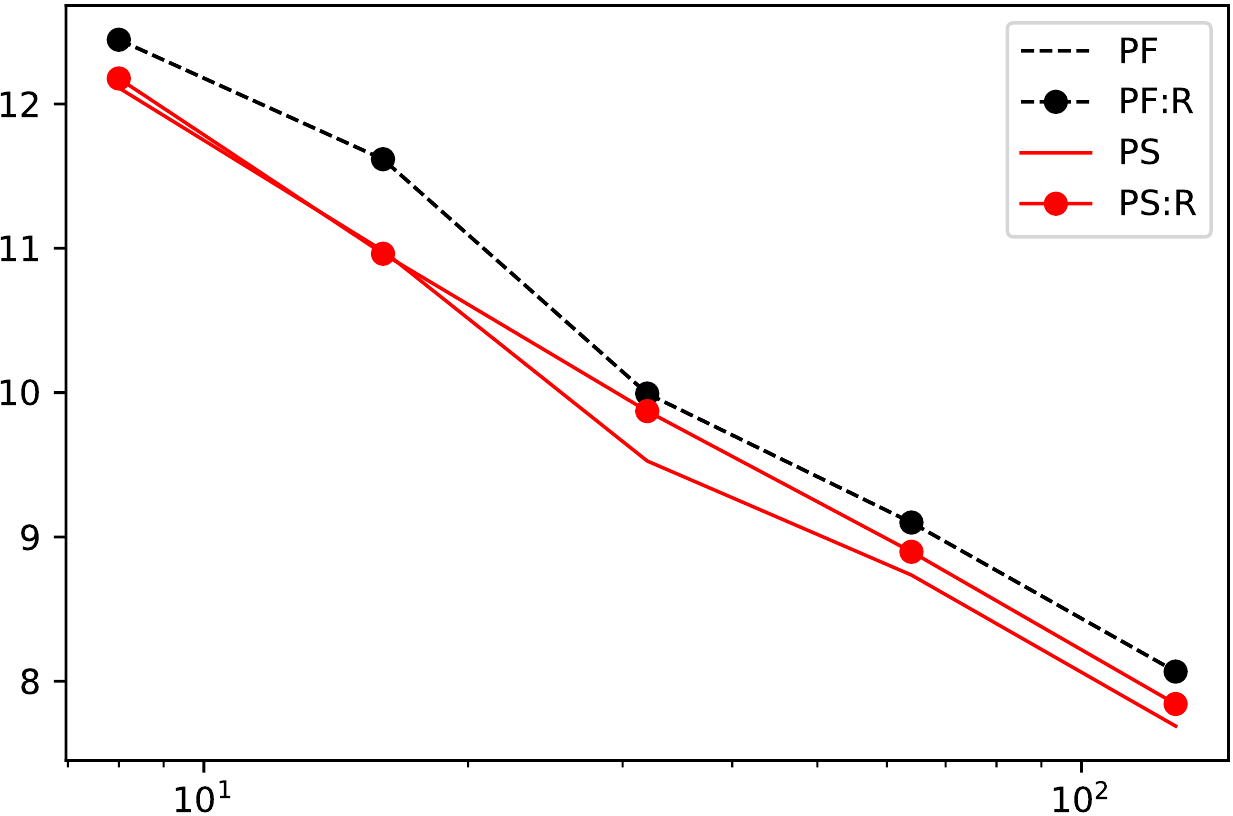}
\caption{source separation: CMUdict}
\label{fig:div-phone}
\end{subfigure}
\caption{Offset KL divergences for the tasks in \cref{sec:tagging,sec:sourcesep}. The logarithmic $x$-axis is the size of particles $M$ ($8 \leq M \leq 128$). The $y$-axis is the offset KL divergence described in \cref{sec:metrics} (in bits per sequence). The smoothing samplers offer considerable speedup: for example, in \cref{fig:divcmu}, the non-resampled smoothing sampler achieves comparable offset KL divergences with only $\nicefrac{1}{4}$ as many particles as its filtering counterparts. Abbreviations in the legend: PF=particle filtering. PS=particle smoothing. BEAM=beam search. `:R' suffixes indicate resampled variants. For readability, beam search results are omitted from \cref{fig:div-phone}, but appear in \cref{fig:cmu-beam} of the appendices.}
\end{figure*}
\section{Experiments}
\label{sec:exps}

In our experiments, we are given a pre-trained scoring model $p_\theta$, and we train the parameters $\phi$
of a particle smoothing algorithm.\footnote{For the details of the training procedures and the specific neural architectures in our models, see \cref{app:training-procedures,app:model-details}.}

We now show that our proposed neural particle smoothing sampler does better than the particle filtering sampler.
To define ``better,'' we evaluate samplers on the \emph{offset KL divergence} from the true posterior.

\subsection{Evaluation metrics}
\label{sec:metrics}
Given \vx, the ``natural'' goal of conditional sampling is for the sample distribution $\phat(\vy)$ to approximate the true distribution $p_\theta(\vy \mid \vx) = \exp G_T / \exp H_0$ from \eqref{eq:condit}.   We will therefore report---averaged over all held-out test examples \vx---the KL divergence
\begin{align}
\text{KL}(\phat || p) &= \E[\vy \sim \phat]{\log \phat(\vy)}  \\ \nonumber
&\qquad - (\E[\vy \sim \phat]{\log \tilde{p}(\vy \mid \vx)} - \log Z(\vx)),
\end{align}
where $\tilde{p}(\vy \mid \vx)$ denotes the {\em unnormalized} distribution given by $\exp G_T$ in \eqref{eq:G}, and $Z(\vx)$ denotes its normalizing constant, $\exp H_0 = \sum_\vy \tilde{p}(\vy \mid \vx)$.

As we are unable to compute $\log Z(\vx)$ in practice, we replace it with
an estimate $z(\vx)$ to obtain an \emph{offset KL divergence}.  This change
of constant does not change the measured difference between two samplers,
$\text{KL}(\phat_1 || p) - \text{KL}(\phat_2 || p)$.
Nonetheless, we try to use a reasonable estimate so that the reported KL
divergence is interpretable in an absolute sense.  Specifically, we take $z(\vx) = \log \sum_{\vy \in {\cal Y}} \tilde{p}(\vy \mid \vx) \leq \log Z$, where ${\cal Y}$ is the full set of distinct particles $\vy$ that we ever drew
for input $\vx$, including samples from the beam search models, while constructing the experimental results graph.\footnote{Thus, $\cal Y$ was collected across all samplings, iterations, and ensemble sizes $M$, in an attempt to make the summation over $\cal Y$ as complete as possible.  For good measure, we added some extra particles: whenever we drew $M$ particles via particle smoothing, we drew an additional $2M$ particles by particle filtering and added them to $\cal Y$.}
Thus, the offset KL divergence
is a ``best effort'' lower bound on the true exclusive KL divergence $\text{KL}(\phat || p)$.

\subsection{Results}

In all experiments we compute the offset KL divergence for both the particle filtering samplers and the particle smoothing samplers, for varying ensemble sizes $M$. We also compare against a beam search baseline that keeps the highest-scoring $M$ particles at each step (scored by $\exp G_t$ with no lookahead). 
 The results are in \cref{fig:div,fig:divcmu,fig:divweibo,fig:div-phone}.

Given a fixed ensemble size, we see the smoothing sampler consistently performs better than the filtering counterpart.  It often achieves comparable performance at a fraction of the ensemble size.\looseness=-1

Beam search on the other hand falls behind on three tasks:  stress prediction and the two source separation tasks. It does perform better than the stochastic methods on the Chinese NER task, but only at small beam sizes. Varying the beam size barely affects performance at all, across all tasks. This suggests that beam search is unable to explore the hypothesis space well.

We experiment with resampling for both the particle filtering sampler and our smoothing sampler. In source separation and stressed syllable prediction, where the right context contains critical information about how viable a particle is, resampling helps particle filtering {\em almost} catch up to particle smoothing.  Particle smoothing itself is not further improved by resampling, presumably because its effective sample size is high.  The goal of resampling is to kill off low-weight particles (which were overproposed) and reallocate their resources to higher-weight ones. But with particle smoothing, there are fewer low-weight particles, so the benefit of resampling may be outweighted by its cost (namely, increased variance).

\section{Related Work}
\label{sec:related}
Much previous work has employed sequential importance sampling for
approximate inference of intractable distributions
\cite[e.g.,][]{thrun99,andrews2017}.
Some of this work learns adaptive proposal distributions in this setting \cite[e.g.][]{gu2015,paige2016}.
The key difference in our work is that we consider future inputs,
which is impossible in online decision settings such as robotics.
\citet{klaas2006} did do particle smoothing, like us, but they did not
learn adaptive proposal distributions.

Just as we use a right-to-left RNN to guide {\em posterior sampling}
of a left-to-right generative model, \citet{krishnan2016} employed a
right-to-left RNN to guide {\em posterior marginal inference} in the
same sort of model.  \citet{serdyuk2017} used a right-to-left RNN to
regularize training of such a model.

\section{Conclusion}

We have described neural particle smoothing, a sequential Monte Carlo
method for approximate sampling from the posterior of incremental
neural scoring models.  Sequential importance sampling has arguably
been underused in the natural language processing community.  It is
quite a plausible strategy for dealing with rich, globally normalized
probability models such as neural models---particularly if a good
sequential proposal distribution can be found.  Our contribution is a
neural proposal distribution, which goes beyond particle filtering in
that it uses a right-to-left recurrent neural network to ``look
ahead'' to future symbols of $\vx$ when proposing each symbol $y_t$.
The form of our distribution is well-motivated.

There are many possible extensions to the work in this paper.
For example, we can learn the generative model and proposal
distribution jointly; we can also infuse them with hand-crafted
structure, or use more deeply stacked architectures; and we can
try training the proposal distribution end-to-end (\cref{fn:end2end}).
Another possible extension would be to allow each step of $q$ to
propose a \emph{sequence} of actions, effectively making the tagset
size $\infty$. This extension relaxes our $|\vy|=|\vx|$ restriction from \cref{sec:intro} and would allow us to do general sequence-to-sequence transduction.

\section*{Acknowledgements}
This work has been generously supported by a Google Faculty Research Award and by Grant No.\@ 1718846 from the National Science Foundation.

\bibliography{psmooth}
\bibliographystyle{acl_natbib}

\clearpage\appendix

\section{The logprob-to-go for HMMs}\label{app:hmm-backward}
As noted in \cref{sec:hmm}, the logprob-to-go $H_t$ can be computed by
the backward algorithm.  By the definition of $H_t$ in \cref{eq:H},
\begin{align}
\exp H_t & = \sum_{\vy_{t:}} \exp\;(G_T - G_t) \label{eq:H-as-diff} \\
         & = \sum_{\vy_{t:}} \exp \sum_{j=t+1}^T g_\theta(\vs_{j-1},x_j,y_j) \\
         & = \sum_{\vy_{t:}} \prod_{j=t+1}^T p_\theta(y_j \mid y_{j-1}) \cdot p_\theta(x_j \mid y_j) \nonumber \\
         & = (\vbeta_t)_{y_t} \text{\ \ (backward prob of $y_t$ at time $t$)} \nonumber
\end{align}
where the vector $\vbeta_t$ is defined by base case $(\vbeta_T)_y = 1$ and for $0 \leq t < T$ by the recurrence
\begin{align}
(\beta_t)_y
&\defeq \sum_{\vy_{t:}} p_\theta(\vx_{t:},\vy_{t:} \mid y_t=y) \label{eq:beta-hmm} \\
&= \sum_{y'} p_\theta(y' \mid y) \cdot p_\theta(x_{t+1} \mid y') \cdot (\vbeta_{t+1})_{y'} \nonumber
\end{align}

The backward algorithm \eqref{eq:beta-oohmm} for OOHMMs in
\cref{sec:oohmm} is a variant of this.

\section{Gradients for Training the Proposal Distribution}\label{app:gradient-derivation}

For a given $\vx$, both forms of KL divergence achieve their minimum of 0 when $(\forall \vy)\; q_\phi(\vy \mid \vx) = p(\vy \mid \vx)$.  However, we are unlikely to be able to find such a $\phi$; the two metrics penalize $q_\phi$ differently for mismatches.  We simplify the notation below by writing $q_\phi(\vy)$ and $p(\vy)$, suppressing the conditioning on $\vx$.

\paragraph{Inclusive KL Divergence}
The inclusive KL divergence has that name because it is finite only when $\textrm{support}(q_\phi) \supseteq \textrm{support}(p)$, i.e., when $q_\phi$ is capable of proposing any string $\vy$ that has positive probability under $p$.  This is required for $q_\phi$ to be a valid proposal distribution for importance sampling.
\begin{align}
\MoveEqLeft[2]	\text{KL} \left( p || q_{\phi} \right)  \\
    &= \E[\vy \sim p]{ \log p\left(\vy  \right) - \log q_\phi(\vy ) } \nonumber \\
 &= \E[\vy \sim p]{\log p \left( \vy  \right)} \nonumber \\
 &\quad - \E[\vy \sim p]{\log q_\phi \left( \vy  \right)} \nonumber
\end{align}
The first term $\E[\vy \sim p]{\log p \left( \vy  \right)}$ is a constant with regard to $\phi$.
As a result, the gradient of the above is just the gradient of the
second term:
\begin{align*}
\nabla_{\phi} \text{KL}(p || q_\phi) &=
\nabla_{\phi} \underbrace{\E[\vy \sim p]{-\log q_\phi \left( \vy  \right)}}_{\text{the cross-entropy }H(p,q_{\phi})}
\end{align*} We cannot directly sample from $p$. However, our weighted mixture $\phat$ from \cref{eq:phat} (obtained by sequential importance sampling) could be a good approximation:
\begin{align}
\nabla_{\phi} \text{KL}(p || q_\phi) &\approx  \nabla_\phi \E[\vy \sim \phat]{-\log q_\phi\left(\vy  \right)}
 \label{eq:cross-entropy}  \\
&= \sum_{t=1}^{T} \E[\phat] {  - \nabla_\phi \log q_\phi(y_t \mid y_{:t-1}, \vx) } \nonumber
\end{align}
Following this approximate gradient downhill has an intuitive interpretation: if a particular $y_t$ value ends up with high relative weight in the final ensemble $\phat$, then we will try to adjust $q_\phi$ so that it would have had a high probability of proposing that $y_t$ value at step $t$ in the first place.

\paragraph{Exclusive KL Divergence}
The exclusive divergence has that name because it is finite only when $\textrm{support}(q_\phi) \subseteq \textrm{support}(p)$.  It is defined by
\begin{align}
\MoveEqLeft[1]
\text{KL}(q_\phi || p) = \E[\vy \sim q_\phi] { \log q_\phi (\vy) - \log p(\vy) } \\
& = \E[\vy \sim q_\phi] { \log q_\phi (\vy) - \log \tilde{p}(\vy) } + \log Z \nonumber \\
&= \sum_{\vy} q_\phi (\vy) \underbrace{\left[ \log q_\phi (\vy) - \log \tilde{p}(\vy) \right]}_{\textrm{call this }d_\phi(\vy)} + \log Z \nonumber
\end{align}
where $p(\vy) = \frac{1}{Z} \tilde{p}(\vy)$ for $\tilde{p}(\vy) = \exp G_T$ and $Z = \sum_{\vy} \tilde{p}(\vy)$.  With some rearrangement, we can write its gradient as an expectation
that can be estimated by sampling from $q_\phi$.\footnote{This is an extension of the REINFORCE
trick \cite{williams92reinforce}, which estimates the gradient of $\E[\vy \sim q_\phi] { \text{reward}(\vy) }$ when the reward is independent of $\phi$.  In our case, the expectation is over a quantity that does depend on $\phi$.}
Observing that $Z$ is constant with respect to $\phi$, first write
\begin{align}\label{eq:exclusive-kl}
\MoveEqLeft[2] \nabla_\phi \text{KL}(q_\phi || p) \\
&= \sum_{\vy} \nabla_\phi \left( q_\phi (\vy) \, d_\phi(\vy) \right) \\
&=  \sum_{\vy} \left( \nabla_\phi q_\phi (\vy)\right) \, d_\phi(\vy)  \nonumber
\\ &\qquad + \sum_{\vy} \underbrace{q_\phi (\vy) \nabla_\phi \log q_\phi (\vy)}_{= \nabla_\phi q_\phi (\vy)} \nonumber \\
&= \sum_{\vy} \left( \nabla_\phi q_\phi (\vy)\right) \, d_\phi(\vy) \nonumber
\end{align}
where the last step uses the fact that $\sum_{\vy}\nabla_\phi q_\phi (\vy) = \nabla_\phi \sum_{\vy} q_\phi (\vy) = \nabla_\phi 1 = 0$.
We can turn this into an expectation with a second use of \newcite{glynn1990}'s observation that
$\nabla_\phi q_\phi (\vy) = q_\phi (\vy) \nabla_\phi \log q_\phi (\vy)$ (the ``likelihood ratio trick''):
\begin{align}
\MoveEqLeft[2] \nabla_\phi \text{KL}(q_\phi || p) \nonumber \\
& = \sum_{\vy}  q_\phi (\vy)  d_\phi(\vy) \nabla_\phi \log q_\phi (\vy) \nonumber \\
& = \E[\vy \sim q_\phi]{d_\phi(\vy) \nabla_\phi \log q_\phi (\vy)} \label{eq:grad} \\
\intertext{which can, if desired, be further rewritten as}
& = \E[\vy \sim q_\phi]{d_\phi(\vy) \nabla_\phi\, d_\phi (\vy)} \nonumber \\
&= \E[\vy \sim q_{\phi}]{\nabla_\phi  \left( \tfrac{1}{2} d_\phi(\vy)^2 \right)} \label{eq:squaredgrad}
\end{align}
If we regard $d_\phi(\vy)$ as a signed error (in the log domain) in trying to
fit $q_\phi$ to $\tilde{p}$, then the above gradient of KL can be interpreted as the
gradient of the mean squared error (divided by 2).\footnote{We thank Hongyuan Mei, Tim Vieira, and Sanjeev Khudanpur for insightful discussions on this derivation.}

We would get the same gradient for any rescaled version of the unnormalized distribution $\tilde{p}$,
but the formula for obtaining that gradient would be different.  In particular, if we rewrite the above derivation but add a constant $b$ to both $\log \tilde{p}(\vy)$ and $\log Z$ throughout (equivalent to adding $b$ to $G_T$), we will
get the slightly generalized expectation formulas
\begin{align}
& \E[\vy \sim q_\phi]{(d_\phi(\vy) - b)\nabla_\phi \log q_\phi (\vy)} \\
& \E[\vy \sim q_{\phi}]{\nabla_\phi  \left( \tfrac{1}{2} \left( d_\phi (\vy) - b \right)^{2} \right)}
\end{align}
in place of \cref{eq:grad,eq:squaredgrad}.  By choosing an appropriate ``baseline'' $b$, we can reduce the variance of the sampling-based estimate of these expectations.
  This is similar to the use of a baseline in the REINFORCE algorithm \cite{williams92reinforce}.
In this work we choose $b$ using an exponential moving average of past $\E{d_\phi (\vy)}$ values:
at the end of each training minibatch, we update $b \leftarrow 0.1\cdot b + 0.9\cdot \bar{d}$, where $\bar{d}$ is the mean of the estimated $\E[\vy \sim q_\phi(\cdot \mid \vx)]{d_\phi (\vy)}$ values for all examples $\vx$ in the minibatch.

\begin{figure}[t]
\centering
\includegraphics[width=\columnwidth]{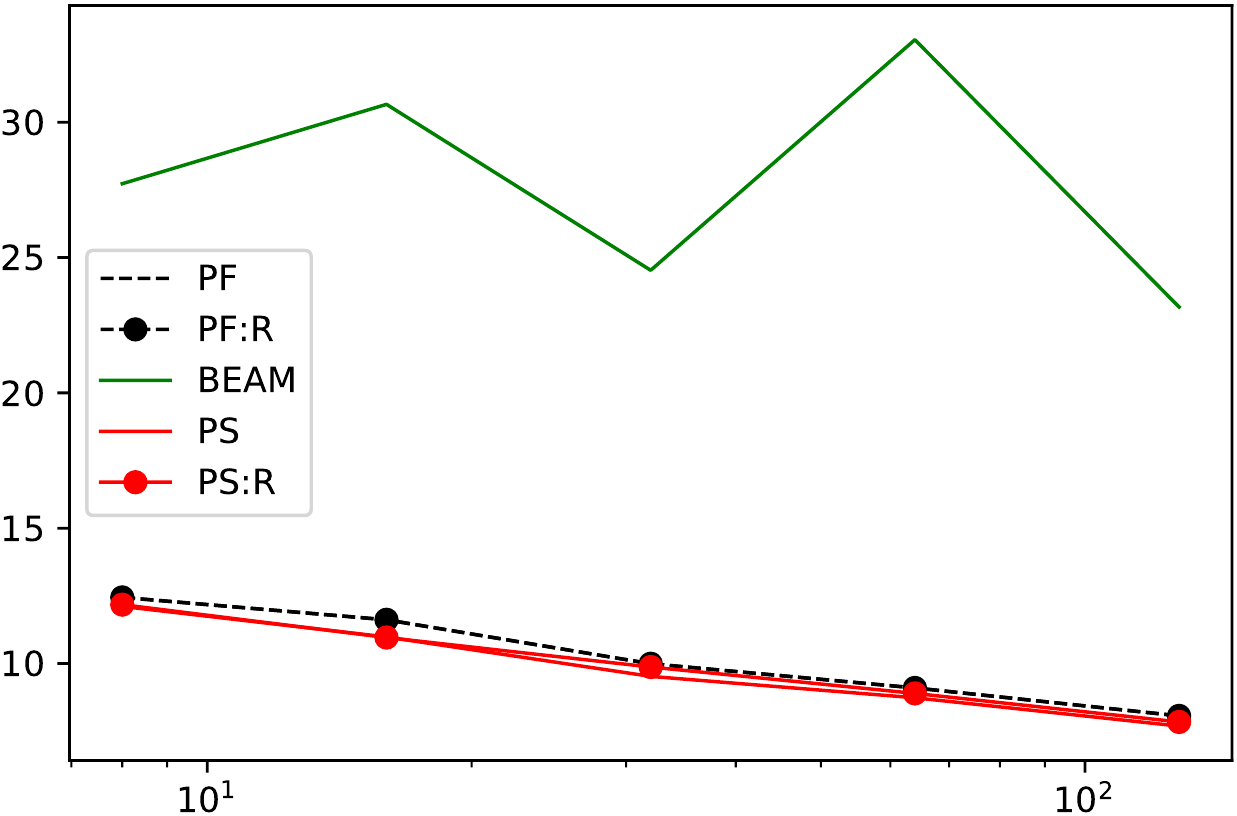}
\caption{Offset KL divergence for the source separation task on phoneme sequences.}
\label{fig:cmu-beam}
\end{figure}

\section{Implementation Details}
\label{app:model-details}
We implement all RNNs in this paper as GRU networks \cite{cho2014} with $d=32$ hidden units (state space $\Real^{32}$). Each of our models (\cref{sec:model}) always specifies the logprob-so-far in \cref{eq:G,eq:s} using a 1-layer left-to-right GRU,\footnote{For the tagging task described in \cref{sec:tagging}, $g_\theta (\vs_{t-1}, x_t, y_t) \defeq \log p_\theta (x_t, y_t \mid \vs_{t-1})$, where the GRU state $\vs_{t-1}$ is used to define a softmax distribution over possible $(x_t, y_t)$ pairs in the same manner as an RNN language model \cite{mikolov2010}.
Likewise, for the source separation task (\cref{sec:sourcesep}), the source language models described in \cref{app:sourcesep-generation} are GRU-based RNN language models.}
while the corresponding proposal distribution (\cref{sec:q}) always specifies the state $\overline{\vs}_t$ in \eqref{eq:sbar} using a 2-layer right-to-left GRU, and specifies the compatibility function $C_t$ in \eqref{eq:C} using a $4$-layer feedforward ReLU network.\footnote{As input to $C_t$, we actually provide not only $\vs_t,\vsbar_t$ but also the states $f_\theta(\vs_{t-1},x_t,y)$ (including $\vs_t$) that could have been reached for {\em each} possible value $y$ of $y_t$.  We have to compute these anyway while constructing the proposal distribution, and we find that it helps performance to include them.}
For the Chinese social media NER task (\cref{sec:tagging}), we use the Chinese character embeddings provided by \citet{peng2015}, while for the source separation tasks (\cref{sec:sourcesep}), we use the 50-dimensional GloVe word embeddings \citep{pennington2014}.  In other cases, we train embeddings along with the rest of the network.  We optimize with the Adam optimizer using the default parameters \citep{kingma2014} and $L_2$ regularization coefficient of $10^{-5}$.

\section{Training Procedures}
\label{app:training-procedures}
In all our experiments, we train the incremental scoring models (the tagging and source separation models described in \cref{sec:tagging} and \cref{sec:sourcesep}, respectively) on the training dataset $T$.  We do early stopping, using perplexity on a held-out development set $D_1$ to choose the number of epochs to train (maximum of $3$).

Having obtained these model parameters $\theta$, we train our proposal distributions $q_{\theta,\phi}$ on $T$, keeping $\theta$ fixed and only tuning $\phi$. Again we use early stopping, using the KL divergence from \cref{sec:metrics} on a separate development set $D_2$ to choose the number of epochs to train (maximum of $20$ for the two tagging tasks and source separation on the PTB dataset, and maximum of $50$ for source separation on the phoneme sequence dataset). We then evaluate $q_{\theta^{*},\phi^{*}}$ on the test dataset $E$.

\bigskip
\bigskip
\noindent\textbf{[\crefrange{app:apps}{app:sourcesep-generation} appear in the supplementary material file.]}

\clearpage

\section{Applications of Sampling}\label{app:apps}

In this paper, we evaluate our sampling algorithms ``intrinsically'' by how well a sample approximates the model distribution $p_\theta$---rather than ``extrinsically'' by using the samples in some larger method.

That said, \cref{sec:apps} did list some larger methods that make use of sampling.  We review them here for the interested reader.

\emph{Minimum-risk decoding} seeks the output
\begin{equation}\label{eq:risk}
\argmin_\vz \sum_\vy p_\theta(\vy \mid \vx) \cdot \text{loss}(\vz \mid \vy)
\end{equation}
In the special case where $\text{loss}(\vz \mid \vy)$ simply asks whether $\vz \neq \vy$, this simply returns the ``Viterbi'' sequence $\vy$ that maximises $p_\theta(\vy \mid \vx)$.  However, it may give a different answer if the loss function gives partial credit (when $\vz \approx \vy$), or if the space of outputs $\vz$ is simply coarser than the space of taggings $\vy$---for example, if there are many action sequences $\vy$ that could build the same output structure $\vz$.  In these cases, the optimal $\vz$ may win due to the combined support of many suboptimal $\vy$ values, and so finding the optimal $\vy$ (the Viterbi sequence) is not enough to determine the optimal $\vz$.

The risk objective \eqref{eq:risk} is a expensive expectation under the distribution $p_\theta(\vy \mid \vx)$.  To approximate it, one can replace $p_\theta(\vy \mid \vx)$ with an approximation $\phat(\vy)$ that has small support so that the summation is efficient.  Particle smoothing returns such a $\phat$---a non-uniform distribution \eqref{eq:phat} over $M$ particles.  Since those particles are randomly drawn, $\phat$ is itself stochastic, but $\E{\phat(\vy)} \approx p_\theta(\vy \mid \vx)$, with the approximation improving with the quality of the proposal distribution (which is the focus of this paper) and with $M$.

In \emph{supervised} training of the model \eqref{eq:condit} by maximizing conditional log-likelihood, the
gradient of $\log p(\vy^*\mid\vx)$ on a single training example $(\vx,\vy^*)$ is
$\nabla_\theta \log p_\theta(\vy^* \mid \vx) = \nabla_\theta G_T^* - \sum_{\vy} p_\theta(\vy \mid \vx) \cdot \nabla_\theta G_T$.  The sum is again an expectation that can be estimated by using $\phat$.  Since $\E{\phat(\vy)} \approx p_\theta(\vy \mid \vx)$, this yields a stochastic estimate of the gradient that can be used in the stochastic gradient ascent algorithm \cite{RobMon51Stochastic}.\footnote{Notice that the gradient takes this ``difficult'' form only because the model is globally normalized.  If we were training a locally normalized conditional model \cite{mccallum-freitag-pereira-2000}, or a locally normalized joint model like \cref{eq:noindep}, then sampling methods would not be needed, because the gradient of the (conditional or joint) log-likelihood would decompose into $T$ ``easy'' summands that each involve an expectation over the small set of $y_t$ values for some $t$, rather than over the exponentially larger set of strings $\vy$.  However, this simplification goes away outside the fully supervised case, as the next paragraph discusses.}

In \emph{unsupervised or semi-supervised training} of a generative model $p_\theta(\vx,\vy)$, one has some training examples where $\vy^*$ is unobserved or observed incompletely (e.g., perhaps only $\vz$ is observed).  The Monte Carlo EM algorithm for estimating $\theta$ \cite{wei1990} replaces the missing $\vy^*$ with samples from $p_\theta(\vy \mid \vx, \text{partial observation})$ (this is the Monte Carlo ``E step''). This \emph{multiple imputation} procedure has other uses as well in statistical analysis with missing data \cite{little-rubin-1987}.

\emph{Modular architectures} provide another use for sampling.  If $p_\theta(\vy \mid \vx)$ is just one stage in an NLP annotation \emph{pipeline}, \newcite{finkel2006} recommend passing a diverse sample of $\vy$ values on to the next stage, where they can be further annotated and rescored or rejected.
More generally, in a \emph{graphical model} that relates multiple strings
  \cite{bouchard-EtAl:2007:EMNLP-CoNLL2007,dreyer-eisner-2009,E17-2120}, inference could be performed by particle belief propagation \cite{ihler-mcallester-2009,lienart-teh-doucet-2015}, or with the help of stochastic-inverse proposal distributions \cite{stuhlmuller-et-al-2013}.  These methods call conditional sampling as a subroutine.

\clearpage
\section{Effect of different objective functions on lookahead optimization}
\label{app:lastchar}
\begin{figure}
\includegraphics[width=\columnwidth]{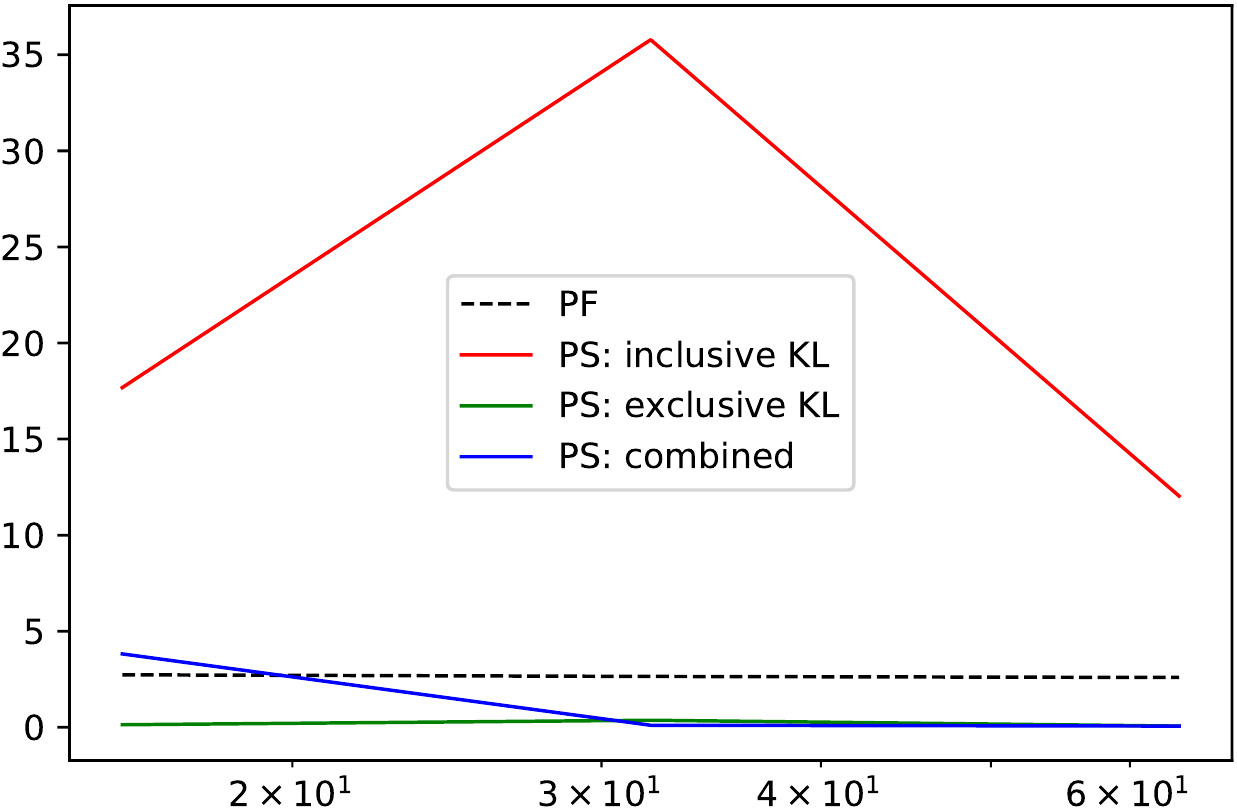}
\caption{Offset KL divergence on the \emph{last char} task: a pathological case where a naive particle filtering sampler
does really horribly, and an ill-trained smoothing sampler even worse. The logarithmic $x$-axis is the particle size used to train the sampler. At test time we evaluate with the same particle size ($M=32$).}
\label{fig:lastchar}
\end{figure}
\Cref{sec:optimization} discussed inclusive and exclusive KL divergences, and gave our rationale for optimizing an interpolation of the two. Here we study the effect of the interpolation weight. We train the lookahead sampler, and the joint language model, on a toy problem called ``last char,'' where $\vy$ is a deterministic function of $\vx$: either a lowercased version of $\vx$, or an identical copy of $\vx$, depending on whether the last character of $\vx$ is \texttt{0} or \texttt{1}.  Note that this problem requires lookahead.

We obtain our $\vx$ sequences by taking the phoneme sequence data from the stressed syllable tagging task and flipping a fair coin to decide whether to append $0$ or $1$ to each sequence.  Thus, the dataset may include $(\vx, \vy)$ pairs such as $(\texttt{K AU CH 0},\;\texttt{k au ch 1})$ or $(\texttt{K AU CH 1},\; \texttt{K AU CH 1})$, but not $(\texttt{K AU CH 1},\;\texttt{k au ch 1})$.

We treat this as a tagging problem, and treat it with our tagging model in \cref{sec:tagging}.
Results are in \cref{fig:lastchar}. We see that optimizing for $\textrm{KL}(\phat||q)$ at a low particle size gives much worse performance than other methods. On the other hand, the objective function $\textrm{KL}(q||p)$ achieves constantly good performance. The middle ground $\frac{\textrm{KL}(\phat||q)+\textrm{KL}(q||p)}{2}$ improves when the particle size increases, and achieves better results than $\textrm{KL}(q||p)$ at larger particle sizes.

\newpage
\section{Generative process for source separation}
\label{app:sourcesep-generation}
Given an alphabet $\Sigma$, $J$ strings $\vx\psup{1}, \vx\psup{2}, \ldots, \vx\psup{J} \in \Sigma^*$ are independently sampled from the respective distributions $p\psup{1}, \ldots p\psup{J}$ over $\Sigma^*$ (possibly all the same distribution $p\psup{1}=\cdots=p\psup{J}$).
These source strings are then combined into a single observed string $\vx$, of length $K=\sum_j K_j$, according to an \defn{interleaving string} $\vy$, also of length $K$.  For example, $\vy=1132123$ means to take two characters from $\vx\psup{1}$, then a character from $\vx\psup{3}$, then a character from $\vx\psup{2}$, etc.  Formally speaking, $\vy$ is an element of the mix language ${\cal Y}_{\vx} = \textsc{mix}(1^{k_1},2^{k_2},\ldots,j^{k_j})$, and we construct $\vx$ by specifying the character $x_k \in \Sigma$ to be $x\psup{y_k}_{|\{i \leq k: y_i=y_k\}|}$.  We assume that $\vy$ is drawn from some distribution over ${\cal Y}_{\vx}$.  The source separation problem is to recover the interleaving string $\vy$ from the interleaved string $\vx$.

We assume that each source model $p\psup{j}(\vx\psup{j})$ is an RNN language model---that is, a locally normalized state machine that successively generates each character of $\vx\psup{j}$ given its left context.  Thus, each source model is in some state $\vs_t\psup{j}$ after generating the prefix $\vx_{:t}\psup{j}$.
In the remainder of this paragraph, we suppress the superscript $\psup{j}$ for simplicity.
The model now stochastically generates character $x_{t+1}$ with probability $p(x_{t+1} \mid \vs_t)$,
and from $\vs_t$ and this $x_{t+1}$ it deterministically computes its new state $\vs_{t+1}$.
If $x_{t+1}$ is a special ``end-of-sequence'' character \eos, we return $\vx=\vx_{:t}$.

Given only $\vx$ of length $T$, we see that $\vy$ could be any element of $\{1,2,\ldots,J\}^T$.  We can write the posterior probability of a given $\vy$ (by Bayes' Theorem) as
\begin{align}
p(\vy \mid \vx) \;\propto\; p(\vy) \prod_{j=1}^{J} p\psup{j}\left(\vx\psup{j}\right)
\label{eq:perm}
\end{align}
where (for this given $\vy$) $\vx\psup{j}$ denotes the subsequence of $\vx$ at indices $k$ such that $y_k=j$.
In our experiments, we assume that $y$ was drawn uniformly from $\cal Y_{\vx}$, so $p(\vy)$
is constant and can be ignored.  In general, the set of possible interleavings ${\cal Y}_{\vx}$ is so large that computing the constant of proportionality (partition function) for a given $\vx$ becomes prohibitive.

\end{document}